%% file: main.tex
\documentclass[sigconf, nonacm]{acmart}

\newcommand\vldbpagestyle{empty}
\newcommand\vldbdoi{10.14778/3570690.3570697}
\newcommand\vldbpages{470 - 479}
\newcommand\vldbvolume{16}
\newcommand\vldbissue{3}
\newcommand\vldbyear{2022}
\newcommand\vldbtitle{\shorttitle} 
\newcommand\vldbavailabilityurl{https://github.com/PKU-DAIR/Hetu/tree/main/tools/Galvatron}

\usepackage{bm}
\usepackage[ruled]{algorithm2e}
\usepackage{multirow}
\usepackage{tikz}
\usepackage{url}
\newcommand*{\circled}[1]{\lower.7ex\hbox{\tikz\draw (0pt, 0pt)%
    circle (.5em) node {\makebox[1em][c]{\small #1}};}}

\usepackage{ulem}

\usepackage{enumerate}

\usepackage{subcaption}
\usepackage{lipsum}
\usepackage{pgfplots}
\usepackage{xcolor}

\usepackage[utf8]{inputenc} 
\usepackage[T1]{fontenc}    
\usepackage{booktabs}       
\usepackage{amsfonts}       
\usepackage{nicefrac}       
\usepackage{microtype}      
\usepackage[utf8]{inputenc} 
\usepackage[T1]{fontenc}    
\usepackage{graphicx}
\usepackage{balance}  
\usepackage{soul}
\usepackage{bbm}
\usepackage[utf8]{inputenc}
\usepackage{amsmath}

\usepackage{amssymb}
\usepackage{amsthm}
\usepackage{booktabs}
\usepackage{amssymb}
\usepackage{makecell}
\usepackage{wrapfig}
\usepackage{hyperref}
\usepackage{enumitem}
\usepackage{float}
\usepackage{footnote}
\usepackage[textsize=small]{todonotes}

\newcommand\blfootnote[1]{%
  \begingroup
  \renewcommand\thefootnote{}\footnote{#1}%
  \addtocounter{footnote}{-1}%
  \endgroup
}

\usepackage{xspace}
\newcommand{\name}{Galvatron\xspace}
\newcommand{\sname}{Galvatron\xspace}

\usepackage{amsmath}

\begin{document}

\title{\name: Efficient Transformer Training over Multiple GPUs Using Automatic Parallelism}

\author{Xupeng Miao$^{*\dagger\ddagger},$~Yujie Wang$^{*\dagger},$~Youhe Jiang$^{*\dagger},$~Chunan Shi$^{\dagger},$~Xiaonan Nie$^{\dagger},$~Hailin Zhang$^{\dagger},$~Bin Cui$^{\dagger\mathsection}$}

\newcommand\vldbauthors{Xupeng Miao, Yujie Wang, Youhe Jiang, Chunan Shi, Xiaonan Nie, Hailin Zhang, Bin Cui}

\affiliation{
\institution{
$^{\dagger}$School of Computer Science \& Key Lab of High Confidence Software Technologies (MOE), Peking University\\
$^{\ddagger}$Carnegie Mellon University\quad $^{\mathsection}$Institute of Computational Social Science, Peking University (Qingdao) \\
\{xupeng.miao, alfredwang, spirited\_away, xiaonan.nie, z.hl, bin.cui\}@pku.edu.cn, youhejiang@gmail.com}
}

\begin{abstract}
Transformer models have achieved state-of-the-art performance on various domains of applications and gradually becomes the foundations of the advanced large deep learning (DL) models. However, how to train these models over multiple GPUs efficiently is still challenging due to a large number of parallelism choices. Existing DL systems either rely on manual efforts to make distributed training plans or apply parallelism combinations within a very limited search space. In this approach, we propose \name, a new system framework that incorporates multiple popular parallelism dimensions and automatically finds the most efficient hybrid parallelism strategy. To better explore such a rarely huge search space, we 1) involve a decision tree to make decomposition and pruning based on some reasonable intuitions, and then 2) design a dynamic programming search algorithm to generate the optimal plan. Evaluations on four representative Transformer workloads show that \name could perform automatically distributed training with different GPU memory budgets. Among all evaluated scenarios, \name always achieves superior system throughput compared to previous work with limited parallelism.\blfootnote{$^*$~Equal contribution.}
\end{abstract}

\settopmatter{printfolios=false}
\maketitle

\pagestyle{\vldbpagestyle}
\begingroup\small\noindent\raggedright\textbf{PVLDB Reference Format:}\\
\vldbauthors. \vldbtitle. PVLDB, \vldbvolume(\vldbissue): \vldbpages, \vldbyear.\\
\href{https://doi.org/\vldbdoi}{doi:\vldbdoi}
\endgroup
\begingroup
\renewcommand\thefootnote{}\footnote{\noindent
This work is licensed under the Creative Commons BY-NC-ND 4.0 International License. Visit \url{https://creativecommons.org/licenses/by-nc-nd/4.0/} to view a copy of this license. For any use beyond those covered by this license, obtain permission by emailing \href{mailto:info@vldb.org}{info@vldb.org}. Copyright is held by the owner/author(s). Publication rights licensed to the VLDB Endowment. \\
\raggedright Proceedings of the VLDB Endowment, Vol. \vldbvolume, No. \vldbissue\ %
ISSN 2150-8097. \\
\href{https://doi.org/\vldbdoi}{doi:\vldbdoi} \\
}\addtocounter{footnote}{-1}\endgroup

\ifdefempty{\vldbavailabilityurl}{}{
\vspace{.3cm}
\begingroup\small\noindent\raggedright\textbf{PVLDB Availability Tag:}\\
The source code of this research paper has been made publicly available at \url{\vldbavailabilityurl}.
\endgroup
}

\section{Introduction}
Transformer models have achieved great success in a wide range of deep learning (DL) applications in recent years, such as computer vision (CV)~\cite{DBLP:ViT,wei2022comparative}, natural language processing (NLP)~\cite{DBLP:gpt3, DBLP:conf/nips/transformer,DBLP:journals/dase/XuCDW22}, graph learning~\cite{graphormer,DBLP:journals/dase/PengCX21} and recommendation systems~\cite{DBLP:BERT4REC}. For example, many Transformer variants (e.g., BERT~\cite{DBLP:conf/naacl/BERT}, GPT-2~\cite{gpt2}, T5~\cite{DBLP:googleT5}) are leading the state-of-the-art performance in various NLP tasks such as machine translation and question answering. Transformers are also applicable to image recognition (e.g, ViT~\cite{DBLP:ViT}, Swin Transformer~\cite{DBLP:conf/iccv/swin}) and multimodal tasks (e.g, CLIP~\cite{DBLP:conf/icml/clip},
DALL-E~\cite{DBLP:conf/icml/dalle}). 
Due to their superior performance, Transformers are becoming increasingly important in modern web companies.

Empirical evidence indicates that scaling model parameters is an effective path towards model performance improvements~\cite{DBLP:journals/corr/abs-2001-08361}. For instance, the original Transformer only has millions of model parameters while GPT-2 has 1.5 billion model parameters with superior performance~\cite{gpt2}. 
Such large amounts of model parameters also incur high computational and memory costs.
Transformers often stack multiple Transformer layers on top of one another and each of them mainly consists of the self-attention module and the feed-forward module. Both of them are dense tensor algebras relying on general-purpose graphical processing units (GPUs) for acceleration. 
With the increasing model scales, building and designing Transformers demand more system optimizations, and \textit{how to perform efficient Transformers training} is becoming more challenging.

Distributed DL systems adopt data and model parallelism to improve the training efficiency by utilizing multiple GPU devices. Data parallelism divides the large volume of input data into multiple parts and each device is only responsible for partial data~\cite{Zinkevich2010ParallelizedSG,DBLP:conf/nips/DeanCMCDLMRSTYN12,DBLP:conf/sigmod/MiaoNSYJM021}. It requires each device to store a whole model replica, suffering from large model scales. Model parallelism is a more promising direction that partitions the model from different \textit{parallelism dimensions} and makes each device store a subset of model parameters, such as tensor parallel~\cite{DBLP:conf/sc/megatron} and pipeline parallel~\cite{DBLP:conf/nips/gpipe,DBLP:conf/icml/NarayananPSCZ21,DBLP:conf/sosp/pipedream,DBLP:conf/mlsys/Yang00RAS21}. Various choices of the parallelism strategies lead to distinct memory consumption, communication overheads and execution efficiency.

However, directly applying these techniques to scaling Transformers is facing crucial challenges in both system efficiency and usability. 
Some recent advanced methods have been proposed to automatically find the parallelism strategies through the fine-grained combination of data and model parallelism for individual operators in the model. For example, OptCNN~\cite{DBLP:conf/icml/optcnn}, FlexFlow~\cite{DBLP:conf/mlsys/flexflow, unger2022unity}, Tofu~\cite{DBLP:conf/eurosys/tofu}, and TensorOpt~\cite{DBLP:journals/tpds/tensoropt} consider both data and tensor parallelism and use different search algorithms to optimize the execution plans. PipeDream~\cite{DBLP:conf/sosp/pipedream} and DAPPLE~\cite{DBLP:conf/ppopp/dapple} combine pipeline parallelism with data parallelism to improve the efficiency. Unfortunately, existing approaches only support limited parallelism dimensions (i.e., data parallelism and rare model parallelism dimensions) or rely on strong model and hardware configurations (i.e., expert-designed parallelism strategy) and result in sub-optimal performance in practice. 
To the best of our knowledge, there is few prior work considering the automatic parallelism for large-scale Transformers with a complex search space including multiple parallelism dimensions.

In this approach, we propose \name, a novel automatic parallel training system for Transformer models over multiple GPUs. Our target is to integrate data parallelism with a variety of model parallelism dimensions, provide a rarely larger search space (compared with previous approaches), and find the optimal hybrid parallelism strategies in an efficient manner.
However, such an integration brings an explosive growth of the search space and cannot be directly explored as usual. Therefore, we are interested in the following question: \textit{How can we exploit as many parallelism dimensions as possible and efficiently explore the search space in the meanwhile?}

We study four popular parallelism paradigms in the distributed training of Transformer models, including data parallelism (DP), sharded data parallelism (SDP)~\cite{DBLP:conf/sc/zero,jiang2022osdp}, tensor parallelism (TP) and pipeline parallelism (PP). 
These paradigms have distinct memory consumption and communication overheads and no single paradigm could beat the others on both sides.
The search space of automatic parallelism should include the arbitrary combinations of these basic parallelism paradigms.
Inspired by some key intuitions from our observations and analysis, we first propose a decision-tree structure to decompose the search space and perform pruning on the tree to remove the inefficient combinations. 
To determine the final strategy for each layer in the model, we then propose a dynamic programming search algorithm to utilize the optimal substructure property of this problem.
It is worth mentioning that the cost estimation in \name considers the GPU performance slowdown from computation and communication overlapping, which has been ignored for a long time in previous parallel training approaches.
We provide an implementation of \name over PyTorch and communication primitives provided by NCCL. 
Unlike existing toolbox-like systems (e.g., DeepSpeed~\cite{rasley2020deepspeed}, Megatron~\cite{DBLP:conf/sc/megatron}) relying on users' expertise and significant tuning efforts,
\name's automatic parallelism only requires a few lines' modifications on the original training script. Our evaluation selects four representative Transformers, including NLP models (i.e., BERT and T5) and CV models (i.e., ViT, Swin Transformer). The experimental results show that \name could significantly outperform the four pure parallelisms and existing automatic parallelisms with limited dimensions (i.e., DP+TP and DP+PP) under various device memory budgets.

We summarize our contributions as follows: First, we enlarge the explored dimension of automatic parallelism for Transformer training, and introduce a novel decision-tree abstraction to decompose the large search space. Second, we design a novel parallelism optimization method to automatically find the most efficient hybrid parallelism strategy based on the estimated costs. Finally, we build \name system that supports larger models' training and achieves up to $338\%$ and 55\% throughput speedups compared to state-of-the-art pure and hybrid parallelism methods respectively.

\vspace{-4mm}
\section{Preliminary}
\subsection{Transformer Models}
Transformers are first proposed to solve sequence modeling and transduction problems such as language modeling and machine translation~\cite{DBLP:conf/nips/transformer}. 
The self-attention and point-wise feed-forward modules are the basic components in each Transformer layer.
Most operations are dense algebras like matrix multiplications, resulting in huge computation costs and memory consumption.

\textit{\textbf{Transformers in NLP.}} Different manners of using Transformer layers in NLP incur three mainly Transformer architectures, including encoder-only (for text classification, e.g., BERT and RoBERTa~\cite{DBLP:journals/corr/abs-1907-11692}), decoder-only (for text generation, e.g., GPT-2 and Transformer-XL~\cite{DBLP:conf/acl/DaiYYCLS19}), and encoder-decoder (for sequence-to-sequence tasks, e.g., T5 and BART~\cite{DBLP:conf/acl/LewisLGGMLSZ20}).
They have similar basic model components and some slight differences on the structures. For example, the decoder has an additional self-attention layer compared to the encoder. What's more, the encoder-decoder architecture combines encoders and decoders symmetrically (i.e., the same number of layers) together. These differences bring some distinct system workload characteristics in both computation and memory.

\textit{\textbf{Transformers in CV.}} Transformers are also becoming increasingly attractive in computer vision areas. Vision Transformer (ViT) first replaces the tokens in languages with patches in images and the patches are fed to the encoder for the image classification task.
Standard ViTs have a fixed number of patches and the same hidden dimension across different layers. Swin Transformer proposes a multi-stage hierarchical architecture with a shifted window-based attention to encode multi-scale patches. However, such multi-scale architectures also uneven computation and memory across layers.

\vspace{-4mm}
\subsection{Parallelism in Distributed Training}

\paragraph{\textbf{Data parallelism}} Data parallelism approaches are widely used to scale up the distributed training for large input datasets. It refers to distribute the data samples across multiple workers to compute and synchronize the model updates (e.g., gradients). Each worker should maintain a replica of the model which implies that the model should be fit into the device memory. To alleviate the redundant memory consumption, DeepSpeed ZeRO~\cite{DBLP:conf/sc/zero} (also named by FSDP in FairScale~\cite{baines2021fairscale}) has been proposed to partition the model states instead of replicating them. It is similar to model parallelism but still follows the data parallelism computation process except involving additional communications to share the model states.

\textit{\textbf{Model parallelism.}} Model parallelism divides the model into multiple parts and each worker is only responsible for the computation of the partial model. Due to the complexity of DL model architectures, a variety of model parallelism approaches have been proposed with different model partition techniques.
There are mainly two kinds of paradigms commonly used for large-scale Transformers training, including distributed tensor parallelism (TP) and layer-wise pipeline parallelism (PP).
For example, Megatron-LM~\cite{DBLP:conf/sc/megatron} uses TP, partitions the feed-forward and self-attention modules in Transformers to multiple devices and inserts communication operations (e.g., All-Reduce) to guarantee consistent results. GPipe~\cite{DBLP:conf/nips/gpipe} first proposes PP, treats each model as a sequence of layers and partitions the model into multiple composite layers across the devices. The workers are organized as a pipeline and transfer intermediate results at the partition boundaries between neighboring partitions.

\textit{\textbf{Automatic parallelism.}} Recent approaches propose to integrate both data and model parallelism and search for better distributed training strategies. For example, FlexFlow, OptCNN, Tofu and TensorOpt consider both tensor parallelism and data parallelism. PipeDream and DAPPLE extend pipeline parallelism and enable data parallelism to replicate each pipeline stage. However, these approaches only explore the combination of data parallelism and at most one single model parallelism dimension. Such limited decision spaces cannot generate efficient enough parallelization plan for many workloads. In fact, industrial companies have taken great efforts to explore better parallelism combinations when training large Transformers on their clusters, such as Turing-NLG~\cite{DBLP:journals/corr/abs-2201-11990} from Microsoft and GPT-3~\cite{DBLP:gpt3} from OpenAI. These evidences suggest that it is necessary to design an automatic parallelization system covering as many parallelism decisions as possible, without relying on strong system tuning experience from human experts.

\textit{\textbf{Task parallelism.}} Some approaches involve multiple training tasks simultaneously. For example, Cerebro~\cite{DBLP:journals/pvldb/NakandalaZK20} targets the model selection problem in AutoML scenarios and each task has similar model architecture with individual configurations (e.g, model size, batch size, learning rate).  
Specifically, with the model hopper parallelism (MOP), Cerobo picks some configurations, each of them is assigned to a worker and performs the training on a sub-epoch of data. The assignment will be adjusted periodically based on the evaluation results. This line of approaches is orthogonal to our problem and they ignore the parallel training of single task.

\section{\name Design}
The goal of \name is to automatically search within the composite parallelism space and generate the optimal parallelization plan for the given Transformer model and the distributed environment.
The key challenge comes from the large search space when considering multiple parallelism strategies and making fine-grained decisions for the model parameters. In this section, we first introduce the search space and then describe our detailed solutions.

\vspace{-2mm}
\subsection{Search Space Analysis}

We first take an example environment with two GPUs to better illustrate the large search space, optimization target and necessary constraints. Then we extend the problem to multi-GPU cases.

\begin{figure}[t]
    \centering
    \includegraphics[width=1.0\linewidth]{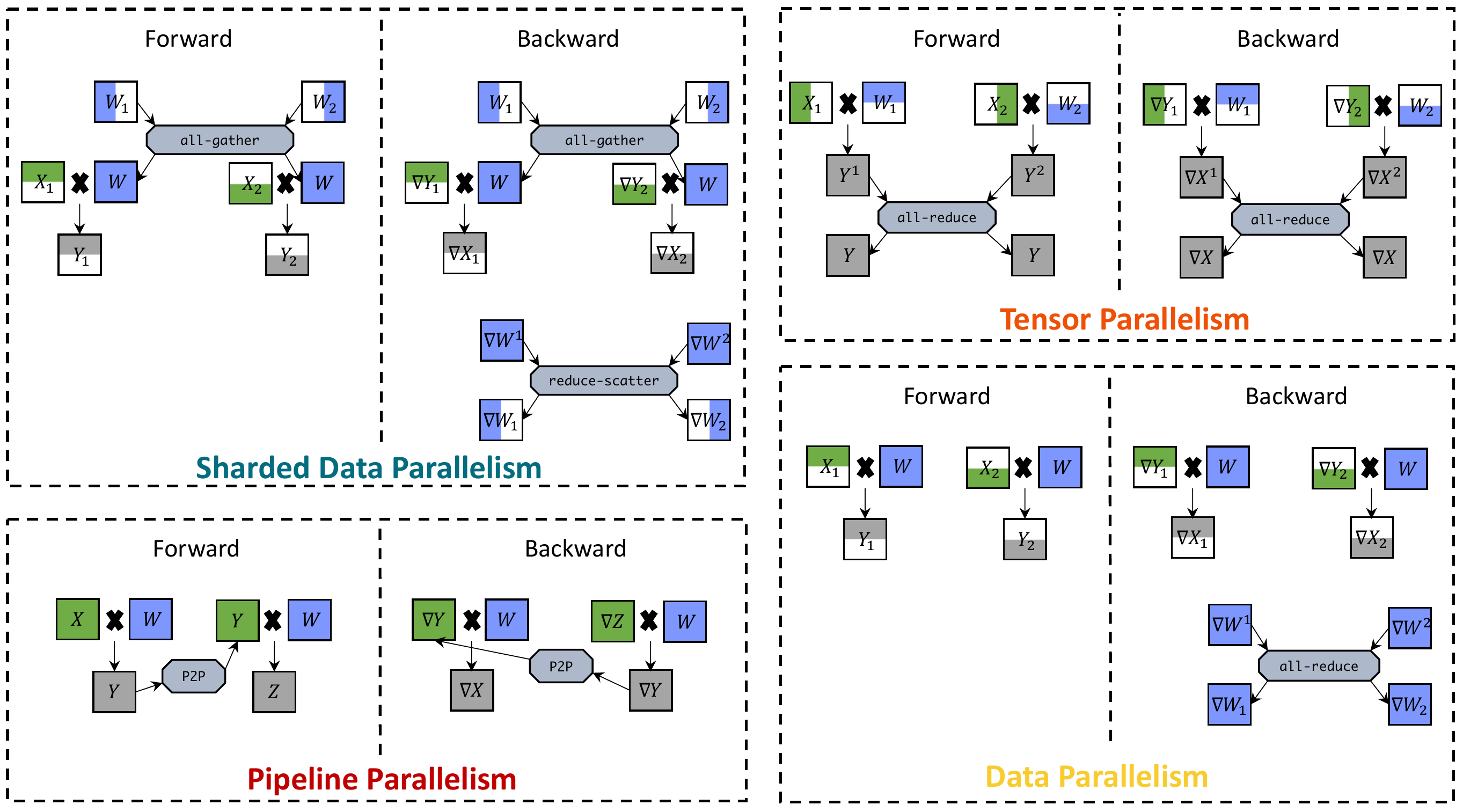}
    \vspace{-6mm}
    \caption{Illustration of different basic parallelisms in \name. We use the green and gray colors to denote the input and output activations for both forward and backward computation. The model parameters and gradients are in blue.}
    \label{fig:parallelism}
    \vspace{-2mm}
\end{figure}

\subsubsection{Two-GPU Example.}
A Transformer model can be treated as a sequence of $L$ layers, and each layer $L_i$ contains a set of model parameters $\mathbf{w}_i$. Due to the back propagation, the forward computation results (i.e., activations) $\mathbf{f}_i$ should be kept inside the device memory before it calculates the gradients $\mathbf{g}_i$ in backward. The problem is to select the optimal parallelism strategy for each layer individually from a large search space, which is a composition of DP, SDP, PP, and TP. As illustrated in Figure~\ref{fig:parallelism}, all these parallelism strategies could split the computation workloads into multiple devices. But they have distinct memory consumption and communication overheads, finally leading to different system efficiency.

\ul{\textit{Data parallelism}}. In DP, each GPU has a model replica and half of the input data samples. Since the size of activations is proportional to the number of data samples, each GPU only needs to store half of the forward activations. After the backward computation, the GPUs should synchronize their gradients (i.e., \texttt{all-reduce}) before updating the model, which has the sample size as model parameters.

\ul{\textit{Sharded Data parallelism}}. In SDP, each GPU has half of model parameters and half of the input data samples. However, it requires two times \texttt{all-gather} to collect the sharded model parameters for forward and backward computation and once \texttt{reduce-scatter} to update gradients. Since an \texttt{all-reduce} operation is equivalent to the combination of once \texttt{all-gather} and once \texttt{reduce-scatter}, the communication cost of SDP is 1.5$\times$ larger than DP.

\begin{figure*}[t]
    \centering
    \includegraphics[width=1.0\linewidth]{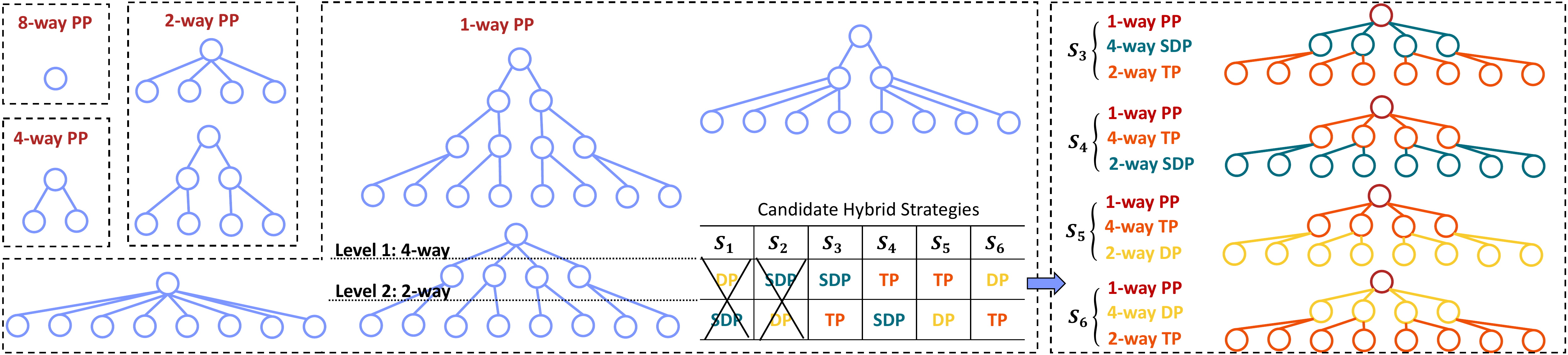}
    \vspace{-6mm}
    \caption{Illustration of the decision trees for 8 GPUs under different PP degrees (i.e., 8/4/2/1). We select one of them to introduce how to use the tree to describe the candidate hybrid parallelism strategies. We remove $S_1$ and $S_2$ as suggested by \textit{Takeaway \#3} and illustrate the left four hybrid strategies on the right part. There are 22 candidate hybrid strategies for all trees in total.}
    \vspace{-4mm}
    \label{fig:tree}
\end{figure*}

\ul{\textit{Pipeline parallelism}}. In PP, the layer $L_i$ could be placed on either GPU 0 or GPU 1, resulting in two possible memory costs: (1, 0) and (0, 1).
The communication cost is mainly determined by whether the neighboring layers are on the same device. 
We select GPipe as the default PP in this approach and the rest (e.g., PipeDream) are left as future work. The efficiency is also affected by the pipeline bubbles (i.e., idle time), which can be reduced by splitting micro-batches.

\ul{\textit{Tensor parallelism}}. In TP, each GPU also has half of model parameters. Unlike SDP, TP allows each device to perform the forward computation (e.g., matrix multiplications and self-attentions) with half model. It requires to synchronize the activations with the \texttt{all-reduce} operations for both forward and backward computation. Due to the intermediate synchronization, TP has some additional replications of the activations.

\subsubsection{Multi-GPU Extension.}
When extending to multi-GPU, the problem becomes more complicated. For example, for two nodes with 4 GPUs in total, it is easy to integrate 2-way TP within a node and 2-way PP across nodes. Alternatively, using 2-way PP within a node and 2-way DP across nodes is also possible. Moreover, there exist hundreds of candidate strategies when scaling to 8 GPUs for a single layer. For a given model, the entire search space is much larger and exponentially growing with the number of layers.

\vspace{-2mm}
\subsection{Decision-tree-based Decomposition}
Considering for such as large search space, it is impossible to brute-force search all the combinations of the four parallelism paradigms within a feasible time budget. Therefore, to explore the search space more efficiently, we introduce the following key intuitions from empirical observations or theoretical analysis.

\ul{\textit{Takeaway \#1}}. PP prefers to be applied across device ``islands''. Each island is a set of devices with higher-bandwidth interconnects (e.g., NVLink, PCIe) and should be in charge of a stage in the pipeline. Compared to other parallelisms, PP has much less communication overheads especially for large models. Because each stage typically has multiple layers but only requires to communicate the activations from the boundary layers. It is sensible to perform PP partition first across slower inter-island links (e.g., QPI, Ethernet).

\ul{\textit{Takeaway \#2}}. Suppose the devices are homogeneous, these parallelism strategies prefer to divide the devices into groups with equal size. For example, a 2-way DP on 4 GPUs means two 2-GPU groups, rather than a single GPU and one 3-GPU group. Consequently, the optimal hybrid parallelism strategy on one group should be also consistent with those of the other groups. Note that, it could fail for PP since the model partitions may have different computation operations, resulting in different optimal parallelism strategies.

Based on the above important intuitions, we design a decision-tree to decompose the search space and represent the candidate hybrid parallelism strategies. We next present the details.

\textit{\textbf{Insights Underpinning Decision-tree.}} We find that most existing automatic parallelism approaches only involve two parallelism dimensions (e.g., OptCNN and FlexFlow), which is easily to enumerate all possible parallelism configurations for a single layer. After involving pipeline parallelism (e.g., PipeDream), they often partition the model into different stages first and each stage is then assigned to a subset of devices. Such kind of observation suggests us to explore the hierarchical search space by utilizing a decision-tree. Another motivation is that we need the tree structure to capture the orders when applying parallelism even inside a stage. Due to the device topology and hierarchical bandwidth, it is necessary to consider the permutations of hybrid strategies since they may have different communication efficiencies.

\textit{\textbf{Decision-tree construction.}}
Given a Transformer model, \name first applies PP to partition the model into multiple stages. In the meanwhile, the devices are also divided into multiple groups with the same size. 
As suggested by Takeaway \#1, it prefers grouping between devices with higher bandwidth.
For an 8-GPU scenario, \name will attempt 1/2/4/8-way PP respectively.
Suppose the model is partitioned evenly by PP, based on Takeaway \#2, the size of the corresponding device group should be 8/4/2/1 respectively after applying PP, which directly determines the number of leaf nodes in our decision-trees.
As shown in Figure~\ref{fig:tree}, given the number of leaf nodes, there might exist multiple possible tree structures. We define the decision-tree construction rules as follows:

\begin{itemize}
    \item Each decision-tree denotes a sub-search-space and its height is the number of available parallelism paradigms. 
    \item Any one of the parallelisms cannot be applied repeatedly in different levels of a decision-tree.
    \item The degree of non-leaf nodes should be selected from $\{2,4,8,\cdots\}$.
\end{itemize}

With the above rules, the constructed trees could represent the arbitrary combinations of these parallelisms in a non-overlap manner.
The guidance from \textit{Takeaway \#1} and \#2 significantly helps \name to avoid the unnecessary and inefficient parallelism combinations.
For a single layer with 8-GPUs, it produces 34 different candidate hybrid parallelism strategies, which reduces the original combinational search space including hundreds of strategies by one order of magnitude. It could be further optimized as follows:

\ul{\textit{Takeaway \#3}}. Using SDP is always better than integrating DP and SDP. We make a comparison with $N$-way DP, $N$-way SDP and the combination of $N_1$-way DP and $N_2$-way SDP ($N_1\times N_2=N$). First, SDP always has fewer model parameters than DP+SDP since $N_2\leq N$. Second, integrating DP and SDP will lead to two rounds of communication including $2(N_1-1)/N_1$ for $N_1$-way DP and $3(N_2-1)/N_2$ for $N_2$-way SDP. Given $N_1\times N_2=N$, we can prove that the minimum value of its cost is still larger than that of pure SDP. Therefore, we exclude such combinations from our search space.
After applying \textit{Takeaway \#3}, we could further reduce the number of candidate strategies to 22 for a single layer with 8-GPUs.

\vspace{-2mm}
\subsection{Parallelism Optimization}

The target of \name is to generate the optimal hybrid parallelism strategy for the input DL model with the given devices.

\textit{\textbf{Problem Formulation.}} We define the optimization problem as follows. Given model $M$ (with $L$ layers) and $N$ devices (with memory capacity of $E$), the object is to find the largest throughput $Tpt$ and return the corresponding parallelism strategy, which is made up of the fine-grained layer-level parallelism strategies.


\begin{algorithm}[t]
\caption{\sname Optimization}
\label{alg:optimizer}
\LinesNumbered 
\KwIn{model: $M$, \#devices: $N$, device memory: $E$}
\KwOut{maximum system throughput $Tpt$}
$Tpt \gets 0$\;
\For{Batch size $B \gets 1, 2, ...$}{
Time costs set $C \gets \{\}$\;
\For{PP degree $P \in \{1, 2, 4, 8, ..., N$)}{ 
Time cost $C_P$ $\gets 0$\;
Model stages $\{M_i\}_{i=1}^{P} \gets \texttt{Pipeline\_Partition}(M, P)$\;
Strategies set $S \gets \texttt{Construct\_Decision\_Tree}(N/P)$\;
\For{$i \in \{1, 2, ..., P\}$}{
$C_P \mathrel{+}= \texttt{Dynamic\_Programming}(E, M_i, B, S)$\;
}
$C$\texttt{.append}($C_P$)\;
}
$C_{opt} \gets \min (C)$\;
\eIf{$C_{opt}$ \textbf{is not} $\infty$}{
    $Tpt \gets \max(B/C_{opt}, Tpt)$\;
}{
    \textbf{return} $Tpt$ \tcc*{Out-Of-Memory}
}
}
\end{algorithm}

\textit{\textbf{Optimization Workflow.}} Basically, the system throughput equals to the ratio between the batch size and the iteration time (i.e., per-batch execution time). 
Tuning the batch size could lead to distinct memory consumption, computation costs and communication overheads.
Scaling the model training with hybrid parallelism strategies could reduce the memory consumption and enlarge the batch size. But it could also bring significant communication overheads. In other words, the highest training throughput does not have to come with the largest batch size.
Therefore, we design the optimization workflow of \name as illustrated in Algorithm~\ref{alg:optimizer}.
It gradually increases the explored batch size (line 2) until exceeding the device memory for all possible parallelism strategies.

Given a candidate batch size $B$, \name then utilizes \textit{Takeaway \#1} to apply PP at first. 
We suppose the total number of devices $N$ is the power of two (e.g., 4, 8, 16), which is common in dedicated GPU training clusters.
So we only explore the 2-th powered PP degrees (line 4).
With a $P$-way PP, the model is evenly partitioned into $P$ stages (line 6). 
Note that, we support several load balancing guidelines for PP partitioning, such as the number of layers/parameters, the maximum memory usage and the execution time. It is also possible to co-optimize by repeatedly interacting with the search inside each stage like Unity~\cite{unger2022unity} and Alpa~\cite{DBLP:conf/osdi/ZhengLZZCHWXZXG22}.
All devices are also evenly divided into $P$ groups. Then we can construct the corresponding decision tree that represents the candidate hybrid parallelism strategies composed of DP, SDP and TP. After obtaining the strategies set $S$, we make the dynamic programming search for each model stage $M_i$ to determine how to parallelize each layer in $M_i$ while minimizing the execution time under the limited device memory budget $E$. The search algorithm returns the minimum time cost if not exceeding the device memory, which is then accumulated for all stages (line 9). Here we exclude the boundary layers' activation transferring costs in PP as they are usually quite small. By comparing the results from all possible PP degrees (line 13) and batch sizes, \name obtains the maximum throughput (line 15).

\textit{\textbf{Dynamic Programming Search.}}
For a given model stage including $L$ layers, we suppose the function $C(L, E)$ represents the total execution time of these $L$ layers under the device memory budget $E$.
We define $c(L, S_j)$ to denote the execution time of the $L$-th layer applying $S_j$, one of the parallelism strategies from the candidates $S$. 
Before applying the dynamic programming, we first prove that the problem follows the optimal substructure property. To obtain the minimum execution time $C(L, E)$, we clarify that the solution must contain the sub-problem solution $C(L^{'}, E^{'})$, which represents the minimum execution time for the sub-model, i.e., first $L^{'}$ layers ($L^{'}\leq L$), within a smaller device memory budget $E^{'}$ ($E^{'}\leq E$). This clarification holds because if the optimal solution $C(L, E)$ does not contain a specific $C(L^{'}, E^{'})$, we can always reduce the total execution time by replacing the sub-problem solution to $C(L^{'}, E^{'})$. Due to the linear sequence model structure, the parallelization plan of the first $L^{'}$ layers will not affect the rest $L-L^{'}$ layers given the same memory budget $E-E^{'}$. Therefore, the problem satisfies the optimal substructure property.
During the search process, we start with $C(0, \cdot)=0$ and $C(\cdot, 0)=\infty$, then we can derive the following state transition formula:
\vspace{-2mm}

\begin{equation}
    \small
    C(L, E) = \min_{S_j\in S}\{C(L-1, E-O(L, S_j)) + c(L, S_j) + R(L, S_i, S_j)\},\label{eq:search}
\end{equation}

\noindent where $O(L, S_j)$ is the memory consumption of the $L$-th layer applying $S_j$ and $R(L, S_i, S_j)$ is the transformation cost between the $L$-th layer applying $S_i$ and its former layer applying $S_j$. If two neighboring layers have different parallelism strategies, the former layer's output should be transformed to the required data layout to facilitate the next layer's parallelism. For example, if the former layer uses the combination between 2-way DP and 2-way TP and the current layer attempts to use 4-way DP, a transformation step is necessary to prepare the full model replica and the $1/4$ forward activation at each device for the current layer. During the state transition process, if the memory usage exceeds the budget $E$, the cost function $C$ should return infinity.

\textit{\textbf{Complexity Analysis.}}
The proposed dynamic programming search formula in E.q.~\eqref{eq:search} has a computation complexity of $\mathcal{O}(LE|S|)$.
As we can see, the size of the single layer's decision space is crucial for the entire complexity and our proposed decision-tree significantly reduces the space and makes it feasible. The number of layers $L$ and the memory budget $E$ also affect the complexity. For extreme cases with thousands of layers or huge memory capacity, we can further reduce the complexity by taking coarse-grained explorations, e.g., fusing multiple layers, using large memory granularity.

\input{overall_experimental_results.tex}

\vspace{-2mm}
\subsection{Cost Estimation}
\name provides a cost estimator to estimate the computation and communication costs and memory consumption during the optimization process. Existing approaches mainly adopt two techniques for the estimation, including \textit{profiling} and \textit{simulating}.
In \name, we take advantages from both sides and design a cost model to make the estimations cheap, efficient and accurate.
Specifically, for the memory consumption, we use the shape of a tensor and its data type to calculate its memory. For the computation time, we suppose it could be estimated by the product of the batch size and the per-sample computation time. The latter could be measured by profiling real layer execution time on a single device. Note that, the Transformers are mainly composed by matrix multiplication operations, so the backward computation is usually twice of the forward computation. For the communication time, we can obtain the approximate communication time by using the size of tensor to be transferred divided by the inter-device connection's bandwidth.

With the above computation and communication cost estimations, $c(l,s)$ (i.e., the cost of a given layer $l$ using a specific parallelism strategy $s\in S$) could be calculated by simulating the execution process. It consists of two steps, e.g., forward and backward computation. 
The simulation for the forward computation is simple and directly sums up the computation and communication costs (i.e., \texttt{all-gather} in SDP and \texttt{all-reduce} in TP). However, during the backward process, DP and SDP enable the computation and communication overlapping, which may bring estimation errors. 
A typical choice is to take the maximum value from the computation and communication costs (e.g., PipeDream~\cite{DBLP:conf/sosp/pipedream}).
Existing automatic parallelism approaches barely notice that modern GPUs simultaneously performing compute kernels and communication primitives (e.g., NCCL~\cite{nccl}) lead to slowdown for both sides.
The performance degradation is mainly from the resource contention of thread warps in GPU streaming multiprocessors.
We find that such contention could slow down the computation and communication by 1.3$\times$ in our evaluations, which is consistent with some recent observations~\cite{DBLP:conf/isca/RashidiD0SSNK21}.
By considering the overlapping slowdown, \name makes more accurate estimations and better optimizations.

\input{implementation.tex}

\input{experiment.tex}

\section{Conclusion}
Large-scale Transformer training is becoming increasingly important due to its expensive training costs. Existing data and model parallelism approaches are suffering from the system efficiency problem. To address the problem, we presented \name, a novel automatic parallel Transformer training system over multiple GPUs. Through the carefully designed search space decomposition and exploration algorithm, \name significantly outperforms the state-of-the-art baselines on the training throughput. We hope the open source release of \name will facilitate the future research directions on more challenging scenarios, e.g., heterogeneous environments and large DL models with complex and dynamic structures.

\section*{Acknowledgments}
This work is supported by the National Natural Science Foundation of China (No. 61832001 and U22B2037) and PKU-Tencent joint research Lab. Bin Cui is the corresponding author.

\newpage
\balance
\normalem
\bibliographystyle{ACM-Reference-Format}
\bibliography{my-reference}



\end{document}

%% file: overall_experimental_results.tex
\begin{table*}[t]
\centering
\small
\caption{Comparison with 8 GPUs under different memory constraints. The maximum throughput (samples/s) of each strategy is given, along with the corresponding batch size in the bracket, and OOM denotes Out-Of-Memory.}
\label{tab:overall_results}
\vspace{-2mm}
\scalebox{0.9}{
\begin{tabular}{c|c|cccccccc}
\toprule 
\makecell{Memory} & \makecell{Strategy} & BERT-Huge-32 & BERT-Huge-48 & ViT-Huge-32 & ViT-Huge-48 & T5-Large-32 & T5-Large-48 & Swin-Huge-32 & Swin-Huge-48 \\
\midrule 
8G  & PyTorch DDP (DP)        & OOM  & OOM   & OOM  & OOM & OOM       & OOM   & OOM  & OOM \\ 
    & Megatron (TP)        & OOM  & OOM   & 16.16 (24) & 10.65 (16)  & OOM  & OOM   & 13.47 (24)  & 8.41 (8) \\ 
    & PyTorch GPipe (PP)        & OOM  & OOM   & 20.57 (56) & 16.59 (32) & OOM    & OOM   & 23.61 (40)  & 16.42 (24) \\ 
    & FSDP/ZeRO-3 (SDP)       & 4.65 (8) & OOM   & 33.25 (64) & 15.71 (40) & 5.97 (8)   & OOM   & 24.86 (48)  & 11.92 (32) \\ 
    & DeepSpeed 3D      & 7.79 (8) & OOM   & 30.56 (40) & 14.59 (16) & 8.12 (8)   & OOM    & 26.22 (32)  & 14.27 (16) \\
    & Galvatron (DP+TP)     & OOM  & OOM   & 29.4 (32) & 15.76 (16) & OOM     & OOM   & 26.18 (24) & 14.76 (16) \\ 
    & Galvatron (DP+PP)     & OOM  & OOM   & 31.79 (48) & \textbf{20.93} (24) & \textbf{9.37} (8)   & OOM   & 27.18 (40)  & 17.71 (24) \\ 
    & Galvatron (ours) & \textbf{8.16} (8) & OOM   & \textbf{36.58} (56) & \textbf{20.93} (24) & \textbf{9.37} (8)   & OOM   & \textbf{31.33} (48)  & \textbf{21.64} (32) \\ 
\midrule
12G & PyTorch DDP (DP)        & OOM       & OOM   & 14.22 (16)  & OOM & OOM      & OOM   & OOM & OOM \\ 
    & Megatron (TP)        & 5.72 (8)  & OOM  & 16.71 (48)  & 10.99 (32)  & 5.14 (8)  & OOM   & 13.68 (40)  & 9.62 (24) \\ 
    & PyTorch GPipe (PP)        & 9.22 (8)  & \textbf{6.2} (8) & 25.13 (104) & 16.62 (64) & 9.09 (8)     & \textbf{6.83} (8)   & 26.07 (72) & 19.82 (48) \\ 
    & FSDP/ZeRO-3 (SDP)       & 8.91 (16) & 3.15 (8) & 47.41 (112) & 24.24 (72) & 11.26 (16)   & 4.11 (8)   & 37.38 (88) & 21.98 (64) \\
    & DeepSpeed 3D      & 7.79 (8) & 5.35 (8)   & 37.88 (80) & 22.68 (48) & 8.12 (8)   & 5.76 (8)   & 34.14 (72)  & 20.07 (40) \\
    & Galvatron (DP+TP)     & 8.92 (8)  & 5.35 (8)& 42.21 (64) & 17.2 (32) & 9.53 (8)    & OOM   & 37.26 (56)  & 20.18 (32)\\ 
    & Galvatron (DP+PP)     & 9.22 (8)  & \textbf{6.2} (8) & \textbf{50.69} (72) & 24.01 (56) & 11.95 (16)   & \textbf{6.83} (8)  & 35.87 (56) & 21.69 (48)\\ 
    & Galvatron (ours) & \textbf{11.39} (16) & \textbf{6.2} (8) & \textbf{50.69} (72) & \textbf{26.63} (72) & \textbf{14.49} (16)    & \textbf{6.83} (8)  & \textbf{41.69} (64) & \textbf{25.42} (64) \\ 
\midrule
16G & PyTorch DDP (DP)  & 6.39 (8)  & OOM   & 44.40 (64)  & OOM & 7.79 (8)      & OOM   & 28.61 (40) & OOM\\ 
    & Megatron (TP)        & 6.06 (16) & 3.88 (8) & 16.81 (72)  & 11.02 (40)  & 5.14 (8)  & OOM   & 13.83 (56)  & 9.71 (40)\\ 
    & PyTorch GPipe (PP)        & 12.96 (16) & 6.2 (8) & 25.26 (144) & 17.24 (96) & 9.09 (8)    & 6.83 (8)   & 28.23 (104) & 20.11 (64)\\ 
    & FSDP/ZeRO-3 (SDP)       & 12.47 (24) & 6.06 (16) & 59.93 (160) & 32.15 (104) & 14.95 (24)   & 7.16 (16)  & 49.68 (136)  & 26.46 (88)\\ 
    & DeepSpeed 3D      & 8.50 (16) & 5.35 (8)   & 41.67 (128) & 25.45 (72) & 11.52 (16)   & 5.76 (8)   & 37.13 (104)  & 24.12 (64) \\
    & Galvatron (DP+TP)     & 12.59 (16) & 6.19 (8) & 46.02 (88) & 23.97 (48) & 14.52 (16)  & 6.84 (8)  & 44.65 (80)  & 26.51 (48) \\ 
    & Galvatron (DP+PP)     & 13.00 (16) & 6.2  (8) & 54.05 (120) & 28.01 (56) & 14.64 (16)   & 6.83 (8)  & 44.15 (96) &  25.82 (56)\\ 
    & Galvatron (ours) & \textbf{15.05} (24) & \textbf{7.46} (16) & \textbf{63.25} (160) & \textbf{35.74} (104) & \textbf{16.50} (24)     & \textbf{8.36} (16)  & \textbf{54.06} (136) & \textbf{29.21} (72) \\ 
\midrule
20G & PyTorch DDP (DP)        & 11.57 (16) & OOM   & 61.54 (112)  & 17.02 (32) & 14.3 (16)   & 5.43 (8)  & 42.82 (80)  & 11.8 (24)\\ 
    & Megatron (TP)       & 6.06 (16) & 3.88 (8) & 16.11 (88)  & 11.02 (56) & 5.47 (16)     & 3.55 (8)  & 13.84 (72) & 9.79 (48)\\ 
    & PyTorch GPipe (PP)        & 13.52 (24) & 7.05 (16) & 28.64 (192) & 17.96 (128) & 9.53 (16)    & 8.13 (16)  & 29.75 (128)  & 20.73 (88)\\ 
    & FSDP/ZeRO-3 (SDP)       & 17.06 (40) & 7.8 (24) & 63.75 (216) &  38.29 (136) & 17.93 (32)     & 7.16 (16)  & 55.22 (176) & 32.63 (120)\\
    & DeepSpeed 3D      & 8.50 (16) & 5.35 (8)   & 43.36 (168) & 27.82 (104) & 13.14 (24)   & 7.96 (16)   & 40.60 (136)  & 26.09 (96) \\
    & Galvatron (DP+TP)     & 14.65 (24) & 8.05 (16) & 61.54 (112) & 28.69 (72) & 15.35 (24)   & 6.84 (8)  & 54.87 (104)  & 30.59 (72)\\ 
    & Galvatron (DP+PP)     & 15.52 (24) & 8.11 (16) & 61.54 (112) & 34.88 (96) & 17.27 (24)   & \textbf{10.33} (16)  & 50.19 (136)  & 31.62 (80)\\ 
    & Galvatron (ours) & \textbf{18.21} (40) & \textbf{8.95} (24) & \textbf{70.5} (152) & \textbf{41.2} (136) &  \textbf{18.64} (32)      & \textbf{10.33} (16) & \textbf{60.06} (144)  & \textbf{37.75} (120) \\ 
\bottomrule
\end{tabular}}
\vspace{-2mm}
\end{table*}

%% file: implementation.tex
\vspace{-2mm}
\section{Implementation}

\name is an automatic parallel training framework especially for Transformer models (open sourced at~\cite{galvatron}), as a part of a novel distributed DL system Hetu~\cite{scis2022hetu,miao2021het,miao2022hetgmp,DBLP:journals/corr/abs-2203-14685}.
We provide a simple and efficient interface to \name users by making a few lines' modifications on the PyTorch training programs~\cite{DBLP:journals/corr/abs-1912-01703,DBLP:journals/pvldb/LiZVSNLPSVDC20}.

\textit{\textbf{Communication group.}}
We implement all communication primitives with PyTorch NCCL functions.
As \name supports complex hybrid parallelism strategies, there could exist many communication groups among the GPUs in the generated parallelization plan. To avoid the expensive NCCL groups construction overheads, \name maintains a global communication group pool which is created in advance and contains all groups that might be used.

\textit{\textbf{Transformation optimization.}}
We propose an efficient \textit{Slice-Gather} step to perform the transformations automatically between two neighboring layers with different parallelism strategies.
Given the previous layer with strategy A and the current layer with strategy B, the main idea of \textit{Slice-Gather} is to ensure the input activations for the current layer are placed on the devices according to the requirement of strategy B, which has been extensively studied~\cite{xu2021gspmd,DBLP:conf/osdi/ZhengLZZCHWXZXG22}.
There exists some special cases that the \textit{Slice-Gather} step brings no communication costs (e.g., strategy A is 4-way TP and strategy is 4-way DP). \name will automatically recognize such cases and finish the transformation without any overheads.

%% file: experiment.tex
\section{Experiments}
\subsection{Experimental Setups}

In this section, we compare \name with 4 pure parallelism strategies implemented by the state-of-the-art systems including PyTorch DDP~\cite{DBLP:journals/pvldb/LiZVSNLPSVDC20} for DP, Megatron~\cite{DBLP:conf/sc/megatron} for TP, PyTorch GPipe~\cite{pytorch_gpipe} for PP, and FairScale FSDP~\cite{xu2020automatic} (similar to DeepSpeed ZeRO Stage-3~\cite{DBLP:conf/sc/zero}) for SDP. 
We also compare with DeepSpeed 3D which is an expert-designed baseline~\cite{deepspeed3d} integrating DP, TP, and PP globally.
Besides, we further provide two auxiliary versions of Galvatron to verify the training efficiency of previous automatic parallelism approaches with limited parallelism dimensions (i.e., DP+TP and DP+PP). To focus on automatic parallelism, we disable some memory optimizations (e.g., recompute~\cite{DBLP:conf/icde/NieMYC22}) and leave them as our future work.
We select NLP models BERT, T5 as well as CV models ViT, Swin Transformer as our experimental models. The statistics of models are listed in Table~\ref{tab:model_config}.
We select the Adam optimizer and use the English Wikipedia and ImageNet-1K as input datasets for them respectively.
Most experiments are evaluated on a single node equipped with 8 Nvidia RTX TITAN 24 GB GPUs using PCIe 3.0. For PP, we manually tune the number of micro-batches to minimize the bubbles and estimate its costs. All results are averaged over 100 iterations.

\begin{table}[t]
\centering
\small
\caption{Statistics of Models}
\label{tab:model_config}
\vspace{-2mm}
\scalebox{0.72}{
\begin{tabular}{c|ccccc}
\toprule 
\makecell{Model} & Layer Num  & Hidden Size & Param. Num & Acti. Size/sample\\
\midrule 
BERT-Huge-32 & 32  & 1280 & 672M & 3149.39MB\\
BERT-Huge-48 & 48  & 1280 & 987M & 4657.51MB\\
BERT-xHuge & 128 & 2560 & 10.2B & 24210.05MB \\
ViT-Huge-32 & 32  & 1280 & 632M & 646.5MB\\
ViT-Huge-48 & 48  & 1280 & 947M & 968.59MB\\
ViT-xHuge & 128 & 2560 & 10.1B & 5313.9MB \\
T5-Large-32 & 16 Enc.+16 Dec.  & 1024 & 502M & 4119.66MB\\
T5-Large-48 & 24 Enc.+24 Dec.  & 1024 & 737M & 6107.75MB\\
Swin-Huge-32 & 2/2/26/2  & 320/640/1280/2560 & 701M & 726.59MB\\
Swin-Huge-48 & 2/2/42/2  & 320/640/1280/2560 & 1016M & 1016.8MB\\
\bottomrule
\end{tabular}}
\vspace{-4mm}
\end{table}

\subsection{End-to-End Comparison}
\autoref{tab:overall_results} shows the overall system throughput results of different models under different strategies with different memory constraints, along with the corresponding batch size.
As we can see, under different model scales and memory budgets, \name always outperforms all baselines in multiple regards. Surprisingly, on ViT models, \name promotes the overall system throughput by up to 338\%, and achieves a maximum of 55\% acceleration compared with hybrid strategies. Similarly, on the other three models, \name still achieves a maximum of 200\%-334\% and 28\%-52\%
compared with single and hybrid strategies respectively.

We can also find that different models may have different preferences on the parallelism strategies. For example, under different memory budgets, BERT almost always prefers DP+PP among all baselines. Similar observations could be also found on some cases of T5. For ViT and Swin Transformer, the preferences change to SDP when increasing the memory budgets. The reason mainly comes from that NLP models have larger activation while CV models have larger model parameters, thus the latter could benefit more from sharding the model parameters across the GPUs. Here DeepSpeed 3D uses an officially suggested strategy~\cite{deepspeed3d} combining 2-way DP/TP/PP together. Such a fixed strategy outperforms three pure parallelisms but fails to beat SDP in most cases.

Another interesting finding is that the hybrid parallelisms like DP+TP and DP+PP may perform worse than pure SDP (e.g., ViT-Huge-32 with 8G, Swin-Huge-32 with 16G).
It further indicates that existing automatic parallelism approaches focusing on limited model parallelism dimensions are suffering from these limitations.

\input{cost_estimation.tex}

\subsection{Estimation Performance}
\autoref{fig:costestimation} demonstrates the cost estimation errors with and without considering the overlapping slowdown. It can be observed that our estimation results are very close to the real execution costs for all experimental models. The average prediction error is less than 5\%. However, when ignoring the slowdown, the estimations become obviously lower, resulting in an average prediction error of more than 15\%, which compromises the promised efficiency of the generated execution strategy.

\input{search_time.tex}

\subsection{Optimization Efficiency}
The efficiency of our dynamic programming search algorithm varies according to different number of model layers, overall strategies and memory constraints. As shown in \autoref{fig:Search_time} (a), when the number of model layers and memory limit increase linearly, the search time of our algorithm increases linearly as excepted, only hundreds of seconds is required to generate the optimal execution plan, which is acceptable and negligible relative to the extremely long model training time. \autoref{fig:Search_time} (b) demonstrates the impact of total parallelism dimensions on the search time, both DP+TP and DP+PP have a total of 4 alternate strategies on 8 GPUs, while Galvatron has 22 overall candidates. In this case, the search time of DP+TP and DP+PP is consistent and much less than that of Galvatron.

\begin{figure*}
    \centering
    \includegraphics[width=1.0\linewidth]{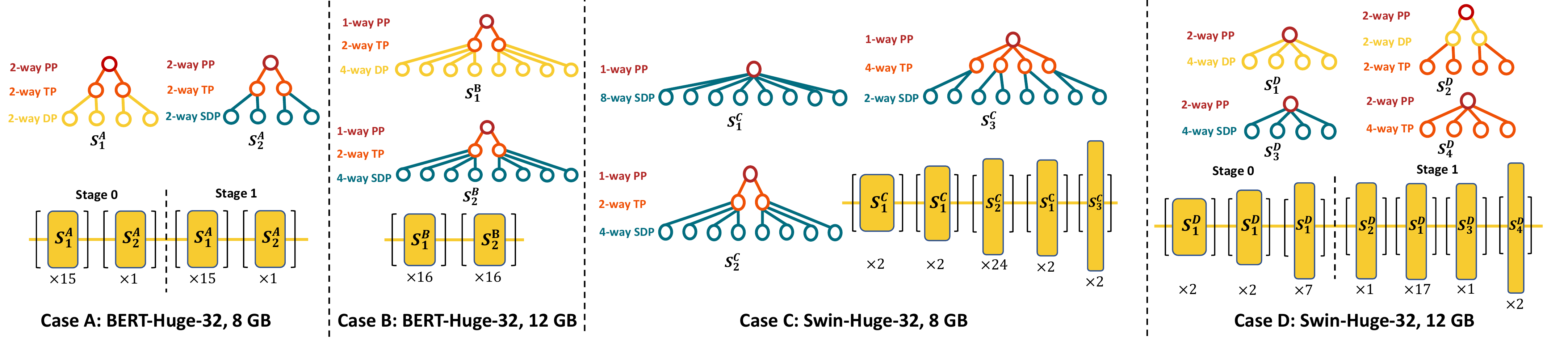}
    \vspace{-8mm}
    \caption{Examples of the optimal parallelism plans given by \name for BERT-Huge-32 and Swin-Huge-32 under 8 GB and 12 GB memory budgets. Each yellow rectangle denotes an encoder layer, and its height and width represent parameter size and activation size respectively. The number $\times N$ under the rectangle means applying an strategy for consecutive $N$ layers.}
    \vspace{-2mm}
    \label{fig:opt_strategy}
\end{figure*}

\begin{table*}[t]
\begin{minipage}{.6\linewidth}
\centering
\small
\caption{Comparison with 16 GPUs.}
\vspace{-2mm}
\label{tab:16gpus_results}
\scalebox{0.9}{
{\begin{tabular}{c|c|cccc}
\toprule 
\makecell{Memory} & \makecell{Strategy} & BERT-Huge-32 & BERT-Huge-48 & ViT-Huge-32 & ViT-Huge-48 \\
\midrule 
8G  & PyTorch DDP (DP)        & OOM  & OOM   & OOM  & OOM \\ 
    & Megatron (TP)        & OOM  & OOM   & 16.86 (32) & 10.86 (16) \\ 
    & PyTorch GPipe (PP)        & 13.79 (16)  & 5.88 (8)   & 50.70 (128) & 27.96 (80) \\ 
    & FSDP/ZeRO-3 (SDP)       & 8.95 (16) & 6.12 (16)   & 69.48 (128) & 34.92 (96) \\ 
    & DeepSpeed 3D & \textbf{15.24} (16) & 6.43 (8) & 57.14 (64) & 29.92 (40) \\ 
    & Galvatron (DP+TP)     & OOM  & OOM   & 54.43 (64) & 24.56 (32) \\ 
    & Galvatron (DP+PP)     & 13.91 (16)  & 5.88 (8)   & 68.56 (128) & 35.02 (72) \\ 
    & Galvatron (ours) & \textbf{15.24} (16) & \textbf{8.43 (16)}   & \textbf{76.74} (128) & \textbf{38.32} (88) \\ 
\midrule 
16G & PyTorch DDP (DP)        & 12.14 (16)  & OOM   & 88.06 (128)  & OOM \\ 
    & Megatron (TP)        & 6.12 (16) & 4.23 (16) & 17.11 (64)  & 11.26 (48)  \\ 
    & PyTorch GPipe (PP)        & 23.29 (40) & 12.92 (24) & 69.72 (320) & 50.23 (208) \\ 
    & FSDP/ZeRO-3 (SDP)       & 30.37 (64) & 11.74 (32) & 123.95 (320) & 61.49 (224) \\ 
    & DeepSpeed 3D & 23.92 (48) & 13.03 (24) & 91.56 (256) & 53.81 (152) \\ 
    & Galvatron (DP+TP)     & 23.01 (32) & 10.50 (16) & 99.22 (160) & 49.82 (96)\\ 
    & Galvatron (DP+PP)     & 23.73 (40) & 13.12  (40) & 115.88 (224) & 61.38 (208)\\ 
    & Galvatron (ours) & \textbf{32.67} (64) & \textbf{14.74} (40) & \textbf{131.15} (320) & \textbf{72.74} (208)\\ 
\bottomrule
\end{tabular}}
}
\end{minipage}\begin{minipage}{.4\linewidth}
\centering
\small
\caption{Comparison with 64 GPUs.}
\vspace{-2mm}
\label{tab:64gpus_results}
\scalebox{0.9}{
{\begin{tabular}{c|c|cc}
\toprule 
\makecell{Memory} & \makecell{Strategy} & BERT-xHuge & ViT-xHuge \\
\midrule 
16G  & PyTorch DDP (DP)        & OOM  & OOM  \\ 
    & Megatron (TP)        & 0.68 (3)  & 1.94 (12)  \\ 
    & PyTorch GPipe (PP)        & 9.74 (16)  & 61.95 (96)   \\ 
    & FSDP/ZeRO-3 (SDP)       & OOM & OOM   \\ 
    & DeepSpeed 3D & 8.44 (16) & 64.91 (96) \\ 
    & Galvatron (DP+TP)     & 1.73 (4)  & 5.07 (2)   \\ 
    & Galvatron (DP+PP)     & 9.74 (16)  & 64.83 (104) \\ 
    & Galvatron (ours) & \textbf{13.77} (24) & \textbf{68.35 (136)}  \\ 
\midrule 
32G  & PyTorch DDP (DP)        & OOM  & OOM  \\ 
    & Megatron (TP)        & 0.77 (7)  & 2.11 (28)  \\ 
    & PyTorch GPipe (PP)        & 21.38 (48)  & 94.84 (288)   \\ 
    & FSDP/ZeRO-3 (SDP)       & OOM & OOM    \\ 
    & DeepSpeed 3D & 21.28 (40) & 91.19 (256) \\ 
    & Galvatron (DP+TP)     & 1.73 (4) & 5.51 (68)   \\ 
    & Galvatron (DP+PP)     & 23.64 (48)  & 110.98 (232) \\ 
    & Galvatron (ours) & \textbf{27.49} (64) & \textbf{114.55 (328)}  \\ 
\bottomrule
\end{tabular}}
}
\end{minipage}
\vspace{-2mm}
\end{table*}

\subsection{Optimal Parallelism Plan}

We list some examples of the optimal parallelism plans suggested by \name. We choose two models, BERT-Huge-32 and Swin-Huge-32, and two memory constraints, 8 GB and 12 GB to analyze. 

For BERT-Huge-32 with 8 GB memory, \name provides an optimal plan containing two strategies $S_1^A$, a combination of PP, TP and DP, and $S_2^A$, a combination of PP, TP, DP. This optimal plan combines all four basic parallelisms, making it possible to train this model within 8 GB memory and achieves a 75\% acceleration compared to other strategies. Under the memory limitation of 12 GB, \name gives a mixture of $S_1^B$, TP+DP, and $S_2^B$, TP+SDP. 
As we can see, \name incorporates SDP and thus reduces memory costs and enlarges the batch size as well as the throughput.

For Swin-Huge-32, the optimal plans given by \name is rather complex, as it has four different layers which have different strategy preference. In Swin Transformer, shallower layers have larger activation size and smaller parameter size.
To reduce memory consumption and communication overhead, shallower layers prefer data parallel which splits input activations and communicates parameter gradients, while deeper layers prefer tensor parallel which splits model parameters and communicates activations. 

\subsection{Scalability Study}

We conduct further comparisons on large clusters. We first extend our experiments to 16 Nvidia RTX TITAN GPUs over two servers connected by 100 Gb InfiniBand network. 
Table~\ref{tab:16gpus_results} illustrates the results on BERT and ViT models. Not surprisingly, Galvatron achieves the best performance with different memory budgets. Compared with the results on 8 GPUs, Galvatron and the hybrid parallelism methods could obtain more than 2$\times$ speedups for many cases. For example, Galvatron enlarges the batch size from 160 to 320 for ViT-Huge-32 under 16 GPUs with 16 GB memory, and the throughput increases from 63.25 to 131.15 samples per second. The 2.07$\times$ speedup comes from the flexible fine-grained layer-level parallelism strategy, which helps to reduce the communication costs and improve the training efficiency.
We also manually search for the optimal DeepSpeed 3D parallelism configurations but they are still unsatisfactory.
We then extend to an industrial GPU cluster including 64 Nvidia A100 GPUs, where each server has 8 GPUs equipped with NVLink and the servers are connected by 100 Gb InfiniBand network. Since the environment scale is significantly larger than before, we also increase the model sizes to 10 billion parameters (i.e., BERT-xHuge and ViT-xHuge, details are in Table~\ref{tab:model_config}). As we can see in Table~\ref{tab:64gpus_results}, even on such a large GPUs cluster, Galvatron still outperforms these baseline methods. Besides, based on our observations, the search time costs do not exponentially grow (i.e., 2.2$\times$ and 9.2$\times$ for 16 GPUs and 64 GPUs respectively compared with 8 GPUs), which is still tolerable. 

%% file: cost_estimation.tex
\begin{figure}[t]
\centering
\begin{minipage}[t]{0.48\linewidth}
\includegraphics[width=1.0\linewidth]{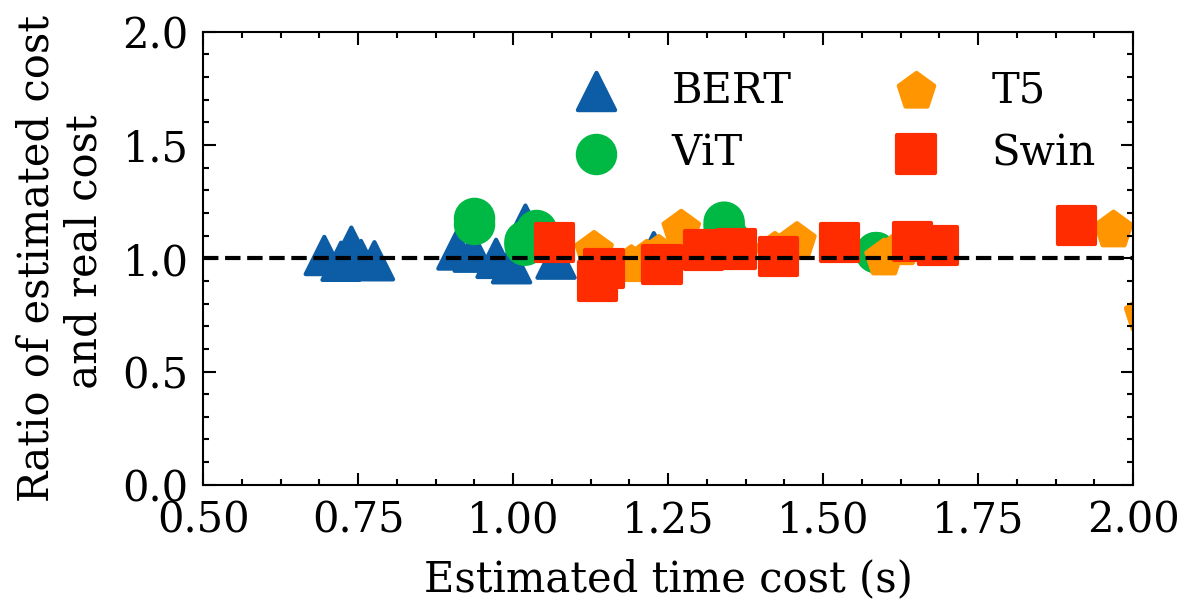}
\vspace{-4mm}
\subcaption{w. overlapping slowdown}
\end{minipage}
\begin{minipage}[t]{0.48\linewidth}
\includegraphics[width=1.0\linewidth]{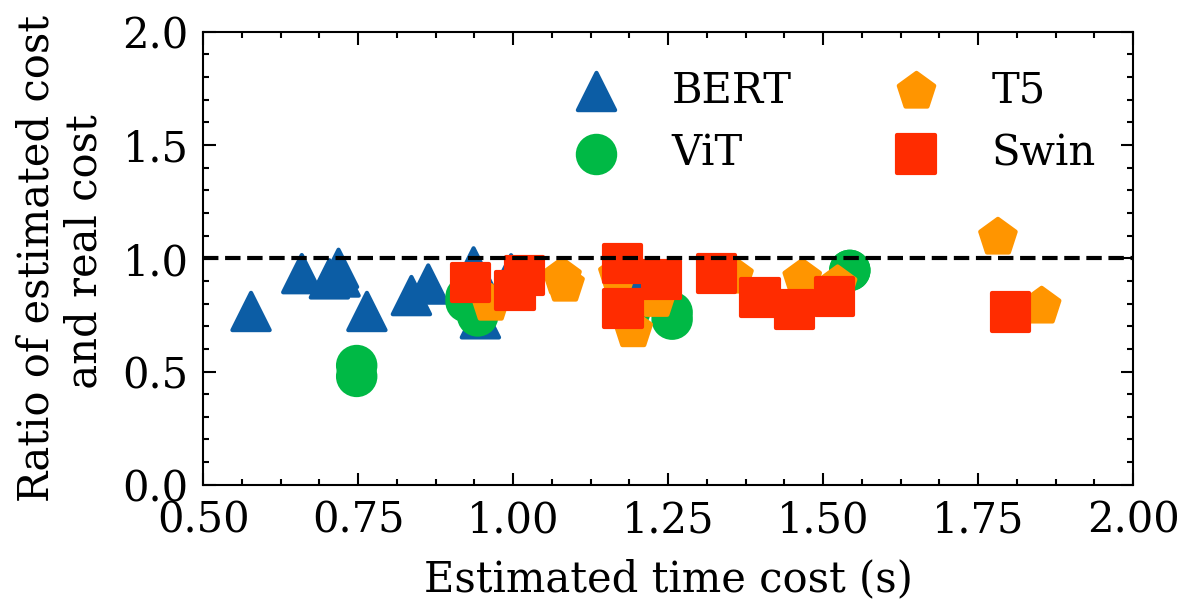}
\vspace{-4mm}
\subcaption{w.o. overlapping slowdown}
\end{minipage}
\vspace{-4mm}
\caption{Estimation errors with and without considering the overlapping slowdown.}\label{fig:costestimation}
\vspace{-2mm}
\end{figure}

%% file: search_time.tex
\begin{figure}
\begin{minipage}[t]{0.23\textwidth}
\begin{tikzpicture}
\small
\begin{axis}[
        legend style={at={(0.8,1.25)},anchor=north, nodes={scale=0.5, transform shape}},
        xlabel={Number of layers},
        ylabel={Time cost (s)}, 
        symbolic x coords={0, 24, 48, 72, 96},
        xticklabels={},
        extra x ticks={0, 24, 48, 72, 96},
        extra x tick labels={0, 24, 48, 72, 96},
        ymajorgrids,
        scale only axis,
        width=0.65\linewidth,
        height=40pt,
        ymin=0,
        ylabel shift=-3pt,
        ylabel near ticks,
        xlabel near ticks,
        xtick pos=bottom,
        y axis line style={opacity=0},
        grid style=dashed,
        legend style={at={(0.3,1.3)},anchor=north, nodes={scale=1.2, transform shape}},
        legend columns=-1,
    ]
    \addplot+[sharp plot] coordinates {
        (0, 0) (24, 31.50363183) (48, 61.8919251) (72, 92.28021836) (96, 122.6685116)
    };
    \addplot+[sharp plot] coordinates {
        (0, 0) (24,  62.47872353) (48, 128.8907599) (72, 195.3027964) (96, 261.7148328)
    };
    \addplot+[sharp plot] coordinates {
        (0, 0) (24, 127.7061207) (48, 253.7028897) (72, 379.6996586) (96, 505.6964276)
    };
    \legend{4G, 8G, 16G}
\end{axis}
\end{tikzpicture}
\vspace{-2mm}
\subcaption{Different memory budgets}
\end{minipage}
\begin{minipage}[t]{0.23\textwidth}
\begin{tikzpicture}
\small
\begin{axis}[
        legend style={at={(0.8,1.25)},anchor=north, nodes={scale=0.5, transform shape}},
        xlabel={Number of layers},
        ylabel={Time cost (s)}, 
        symbolic x coords={0, 24, 48, 72, 96},
        xticklabels={},
        extra x ticks={0, 24, 48, 72, 96},
        extra x tick labels={0, 24, 48, 72, 96},
        ymajorgrids,
        scale only axis,
        width=0.65\linewidth,
        height=40pt,
        ymin=0,
        ylabel shift=-3pt,
        ylabel near ticks,
        xlabel near ticks,
        xtick pos=bottom,
        y axis line style={opacity=0},
        grid style=dashed,
        legend style={at={(0.3,1.3)},anchor=north, nodes={scale=1.2, transform shape}},
        legend columns=-1,
    ]
    \addplot+[sharp plot] coordinates {
        (0, 0) (24, 10.32815886) (48, 20.97259283) (72, 31.61702681) (96, 42.26146078)
    };
    \addplot+[sharp plot] coordinates {
        (0, 0) (24,  10.81789231) (48, 21.54467845) (72, 32.13630366) (96, 46.23788142)
    };
    \addplot+[sharp plot] coordinates {
        (0, 0) (24, 127.7061207) (48, 253.7028897) (72, 379.6996586) (96, 505.6964276)
    };
    \legend{DP+TP, DP+PP, Galvatron}
\end{axis}
\end{tikzpicture}
\vspace{-2mm}
\subcaption{\name vs DP+TP vs DP+PP}
\end{minipage}
\vspace{-4mm}
\caption{Search time costs with different numbers of layers.}\label{fig:Search_time}
\vspace{-4mm}
\end{figure}